\documentclass[10pt,twocolumn,letterpaper]{article}

\usepackage[pagenumbers]{iccv} 

\usepackage{afterpage}

\definecolor{iccvblue}{rgb}{0.21,0.49,0.74}
\usepackage[pagebackref,breaklinks,colorlinks,allcolors=iccvblue]{hyperref}

\def\paperID{6} 
\def\confName{ICCV}
\def\confYear{2025}

\title{PhraseStereo: The First Open-Vocabulary Stereo Image Segmentation Dataset}

\author{Thomas Campagnolo\textsuperscript{1,2} , 
Ezio Malis\textsuperscript{1} , 
Philippe Martinet\textsuperscript{1} ,
Gaetan Bahl\textsuperscript{2}\\
\textsuperscript{1}Centre Inria d'Universite Cote d'Azur, France 
\textsuperscript{2}NXP Semiconductors, France \\
{\tt\small \{thomas.campagnolo, ezio.malis, philippe.martinet\}@inria.fr,} \\
{\tt\small \{thomas.campagnolo, gaetan.bahl\}@nxp.com}
}

\begin{document}
\maketitle
\begin{abstract}
Understanding how natural language phrases correspond to specific regions in images is a key challenge in multimodal
semantic segmentation. 
Recent advances in phrase grounding are largely limited to single-view images, 
neglecting the rich geometric cues available in stereo vision. For this, we introduce PhraseStereo, the first novel dataset that brings 
phrase-region segmentation to stereo image pairs. PhraseStereo builds upon the PhraseCut dataset by leveraging 
GenStereo to generate accurate right-view images from existing single-view data, enabling the extension of phrase 
grounding into the stereo domain. This new setting introduces unique challenges and opportunities for multimodal 
learning, particularly in leveraging depth cues for more precise and context-aware grounding.
By providing stereo image pairs with aligned segmentation masks and phrase annotations, PhraseStereo lays the 
foundation for future research at the intersection of language, vision, and 3D perception, encouraging the 
development of models that can reason jointly over semantics and geometry.
The PhraseStereo dataset will be released online upon acceptance of this work.
\vspace{-10pt}
\end{abstract}    
\section{Introduction}
\label{sec:intro}

Interpreting natural language phrases with the corresponding specific regions in images is a core challenge in 
multimodal AI, for tasks such autonomous robot navigation, human-robot interaction, and more.
While recent advances in phrase grounding have improved performance on single-view images, these methods often 
ignore the rich geometric information available in stereo vision. This limitation restricts their ability to reason 
about spatial relationships and depth, a key aspect of real-world understanding.

Despite the progress in grounding natural language phrases to image regions, existing works are 
predominantly designed for mono RGB images with datasets such as ReferIt~\cite{kazemzadeh-etal-2014-referitgame}, RefCOCO~\cite{yu2016modeling}, Google Referring Expressions~\cite{mao2016generation}, 
and PhraseCut~\cite{wu2020phrasecut}. However, these datasets lack detailed geometric representations.

\begin{figure}[t]
  \centering
  \includegraphics[page=1,width=0.42\textwidth]{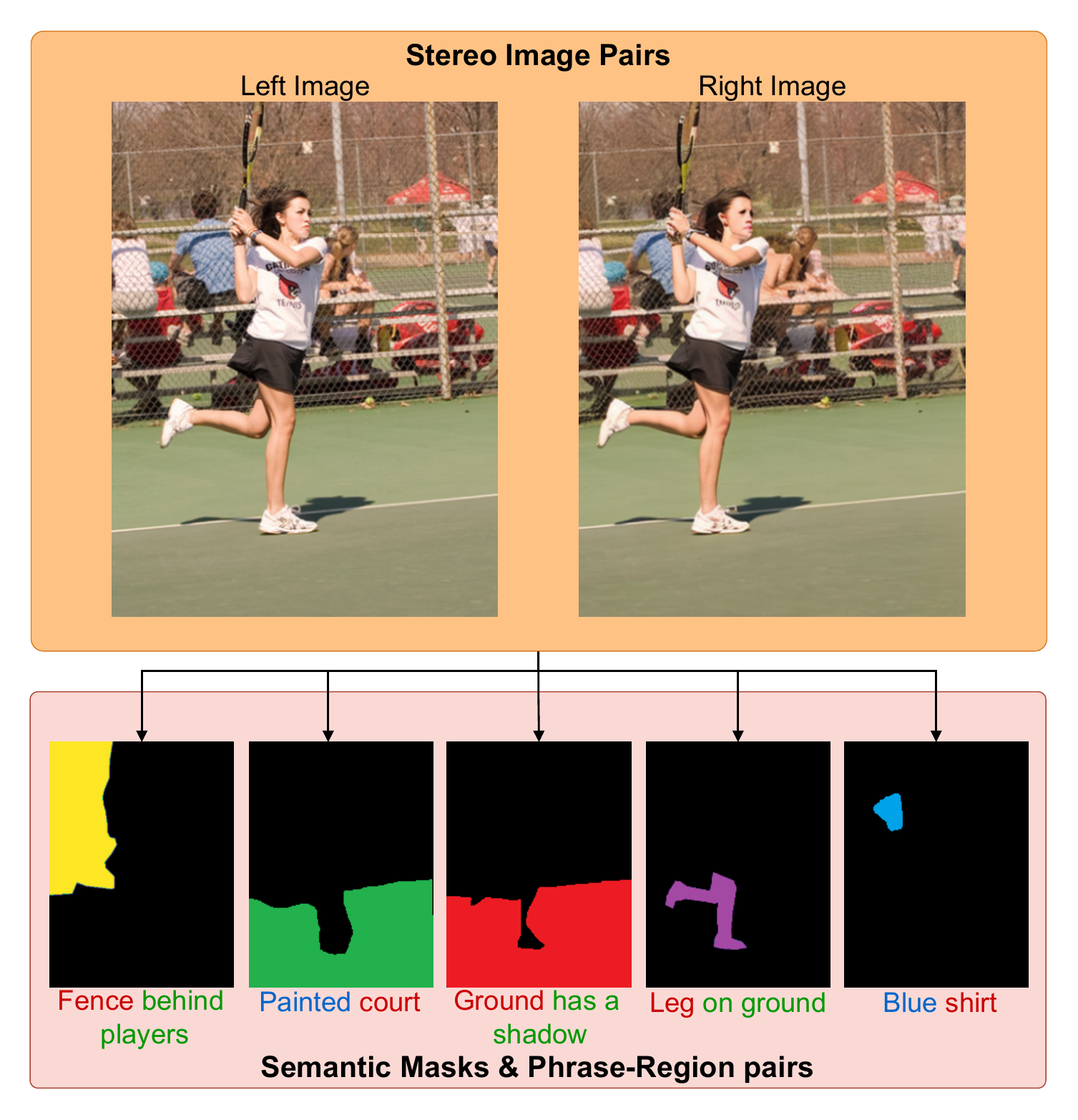}
  \caption{Stereo image pairs with corresponding semantic segmentation masks from PhraseStereo dataset. 
  Each phrase-region pair is color-coded to highlight its linguistic meaning: attributes in blue, categories in red, and relationships in green.}
  \vspace{-10pt}
  \label{fig:phrase_seg_region}
\end{figure}

The main contribution of this paper is the introduction of a new dataset about phrase grounding segmentation with stereo image pairs, 
enabling models to leverage stereo geometry for more accurate segmentation of referred objects and regions.
PhraseStereo dataset most extends PhraseCut~\cite{wu2020phrasecut} to the stereo domain. It focuses on two important aspects: (1) the use of stereo images and  
(2) the consequently geometric spatial relationships between image pairs.
PhraseStereo contains 77,262 stereo images and 345,486 phrase-region annotations, with multiple annotations per image pair, 
as illustrated in \cref{fig:phrase_seg_region}, following the structure and properties of PhraseCut dataset~\cite{wu2020phrasecut}. 
The right-view images are generated using GenStereo~\cite{qiao2025genstereo}, a stereo generative model applied to the original single images. 
PhraseStereo's task involves segmenting regions in stereo image pairs based on natural language phrases, leveraging geometric 
information to improve spatial segmentation and address challenges inherent to single-image approaches, such as occlusions and 
depth ambiguities.

Our contributions are summerized as follows:
\begin{itemize}
  \item We introduce PhraseStereo, the first dataset for open-vocabulary stereo semantic segmentation.
  PhraseStereo extends PhraseCut~\cite{wu2020phrasecut} to the stereo domain, enabling models to exploit 
  geometric context from stereo vision for more precise segmentation of phrase-referred objects and regions.
  \item We address the challenge of hallucinations, relative to the right-view image generation step. To this
  end, we conducted a detailed analysis using perceptual similarity metrics, including SSIM and LPIPS. 
\end{itemize}

\section{Related Work}
\label{sec:relatedWork}

\Cref{tab:related_works} presents a comparison of datasets related to grounding referring expressions.
Kazemzadeh \etal~\cite{kazemzadeh-etal-2014-referitgame} introduced the ReferItGame, a two-player interactive
framework designed to collect the ReferIt dataset. In this setup, one participant generates natural 
language expressions to refer to specific objects within real-world scene images, 
while the other identifies the target object by clicking on its location.
RefCOCO~\cite{yu2016modeling} also uses the ReferItGame, but focuses specifically on appearance-based 
descriptions (e.g., “the man in the yellow polka-dotted shirt”) rather than positional or 
relational cues (e.g., “the second man from the left”), aiming to isolate visual grounding from 
spatial reasoning.
Mao \etal~\cite{mao2016generation} introduced Google RefExp, which builds upon the methodology of 
ReferIt~\cite{kazemzadeh-etal-2014-referitgame} while utilizing the MSCOCO dataset~\cite{lin2014microsoft}. 
Unlike ReferIt, MSCOCO provides instance-level segmentation annotations for 80 predefined object categories.

The SUN-Spot dataset~\cite{mauceri2019sun} provides both RGB images and depth maps to support the localization of 
objects using spatial referring expressions. However, its use of depth is limited to interpreting 
spatial language involving prepositions like \textit{"behind"} and \textit{"in front of"}. 
Chen \etal~\cite{chen2020scanrefer} introduced the ScanRefer dataset, which builds upon ScanNet~\cite{dai2017scannet} 
by incorporating RGB-D scans where depth information is encoded as part of the 3D point cloud representation. 
Designed for 3D object localization from natural language descriptions, ScanRefer is limited to indoor environments, 
potentially restricting its generalization to more diverse scene types.

The PhraseCut dataset~\cite{wu2020phrasecut} is derived from Visual Genome~\cite{krishna2017visual} and grounds natural language phrases to image 
regions. It includes segmentation masks for the regions corresponding to each phrase, 
enabling fine-grained phrase-level supervision. In contrast, PhraseStereo encodes geometric information directly within stereo image pairs.
To our knowledge, PhraseStereo is the only dataset that combines referring expression annotations 
with stereo image-based semantic segmentation, while also offering diverse scenes across both 
indoor and outdoor scene images.

\begin{table}[h]
  \centering
  \resizebox{0.48\textwidth}{!}{%
  \begin{tabular}{@{}lccc@{}}
    Dataset & Data Format & Tasks & Geometric Context \\
    \midrule
    ReferIt~\cite{kazemzadeh-etal-2014-referitgame} & Image & S & No \\
    RefCOCO~\cite{yu2016modeling} & Image & S & No \\
    Google RefExp~\cite{mao2016generation} & Image & S & No \\
    SUN-Spot~\cite{mauceri2019sun} & Image & L & Yes (Depth map) \\
    ScanRefer~\cite{chen2020scanrefer} & Image & L & Yes (3D point cloud) \\
    PhraseCut~\cite{wu2020phrasecut} & Image & S & No \\
    \textbf{PhraseStereo (ours)} & Stereo Images & S & Yes (Stereo encoded) \\
    \bottomrule
  \end{tabular}
  }
  \caption{Comparison between competing datasets. S indicates semantic segmentation tasks, L indicates 
  localization tasks. PhraseStereo uniquely integrates referring expression segmentation 
  with stereo-based 3D geometric vision.}
  \vspace{-10pt}
  \label{tab:related_works}
\end{table}

\section{PhraseStereo dataset}

\begin{figure*}[t]
  \centering
    \includegraphics[page=1,width=0.77\textwidth]{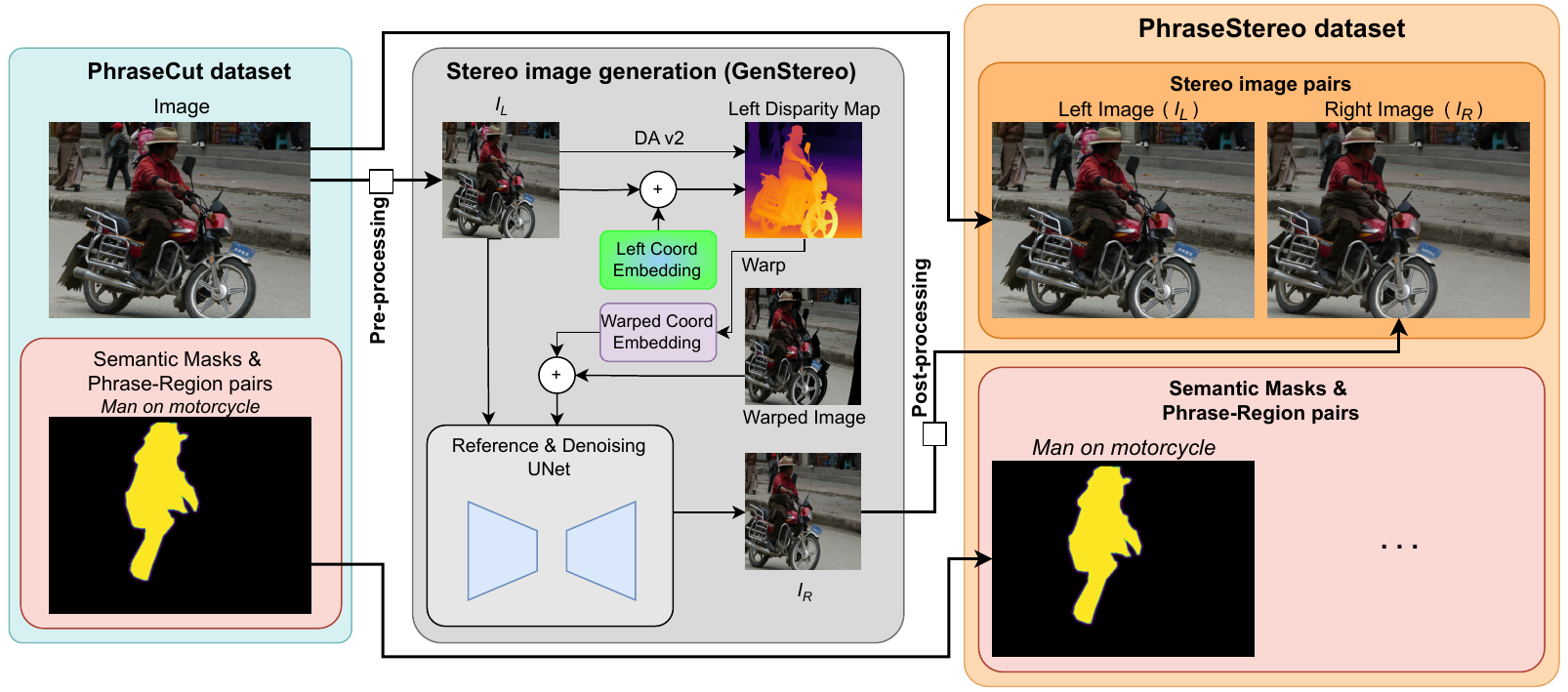}
  \caption{Overview of the PhraseStereo dataset generation pipeline. Starting from a PhraseCut~\cite{wu2020phrasecut}
  image $I_L$ (left), we apply a pre-processing step including image resizing. GenStereo~\cite{qiao2025genstereo}, is then used to generate the corresponding right-view image $I_R$ (center). Post-processing
  includes resizing $I_R$ back to the original resolution of $I_L$, ensuring stereo consistency.
  The resulting stereo pair ($I_L$, $I_R$), along with the phrase-region annotations pairs,
  forms the PhraseStereo dataset (right).}
  \vspace{-10pt}
  \label{fig:pipeline}
\end{figure*}

In this section, we describe the composition of PhraseStereo. Our (left) images and annotations are derived from the PhraseCut~\cite{wu2020phrasecut}, which provides natural 
language phrases and corresponding segmentation masks. 
To extend this into the stereo domain, we generate the right-view images using GenStereo~\cite{qiao2025genstereo}.
The complete data collection pipeline, illustrated in \cref{fig:pipeline}, 
consists of four main stages: image pre-processing, stereo image generation, right image post-processing, 
and data transferring and composition.

\vspace{-10pt}
\paragraph{Image pre-processing}

The images of PhraseCut dataset~\cite{wu2020phrasecut} vary widely in resolution, ranging from small to very high dimensions. 
To ensure consistency and compatibility, in the pre-processing stage we resize the image to
$512\times512$ pixels. 
This resolution offers a practical compromise: it preserves sufficient visual detail while standardizing input size across diverse image resolutions.
Although GenStereo originally employed cropping or, more recently, patching strategies to meet its 
input size requirements~\cite{qiao2025genstereo}, these methods either exclude parts of the image or introduce artifacts 
that compromise the visual integrity of the generated right-view image. 
In contrast, our resizing approach preserves the entire image content, 
ensuring better alignment and visual coherence in the resulting stereo pairs.
\vspace{-10pt}
\paragraph{Stereo image generation}

To generate the right images in PhraseStereo, our pipeline employs GenStereo~\cite{qiao2025genstereo}, a diffusion-based framework for 
stereo image synthesis. Given the (left-view) image from PhraseCut~\cite{wu2020phrasecut}, $I_L$, the goal is to generate 
a high-quality right-view image $I_R$ that preserves both visual fidelity and geometric consistency. 
GenStereo treats the left image as the reference and synthesizes the right image as the target. 
The corresponding disparity $D$ is obtained from a predicted depth map using Depth Anything V2~\cite{yang2024depth}, 
a monocular depth estimation model. This disparity information is used to compute disparity-aware 
coordinate embeddings, which, along with a warped version of the input image, conditions the 
diffusion process for improved stereo alignment. Additionally, GenStereo incorporates an adaptive 
fusion mechanism that blends the diffusion-generated image with the warped input, enhancing both 
realism and disparity consistency in the final stereo pair.
\vspace{-10pt}
\paragraph{Right image post-processing}

After generating the right-view image $I_R$, the output resolution corresponds to 
the resized input image $512\times512$ pixels. To restore the original spatial alignment with the left-view 
image $I_L$, we perform a final resizing step. Knowing the original resolution of $I_L$, we resize 
$I_R$ back to match it, ensuring that the stereo pair maintains consistent dimensions.
\vspace{-10pt}
\paragraph{PhraseStereo composition}

PhraseStereo is composed of the original left-view image $I_L$ from the PhraseCut dataset~\cite{wu2020phrasecut}, the
right-view image $I_R$ generated by GenStereo~\cite{qiao2025genstereo}, and the corresponding phrase-region 
annotations. Our final dataset includes 345,486 referring expressions across 77,262 stereo image 
pairs. Following the original PhraseCut splits, PhraseStereo is divided into 71,746 stereo image 
pairs for training with the corresponding 310,816 phrases, 2,971 stereo image pairs for validation
(with the related 20,316 phrases), and 2,545 stereo image pairs with 14,354 phrases for testing. 
The phrase-region annotations are directly transferred from PhraseCut~\cite{wu2020phrasecut}, preserving the original 
structure.

\section{Results and Analysis}

The PhraseStereo task involves generating a binary segmentation mask for a given input 
image pairs, conditioned on a referring natural language phrase and informed by stereo geometric cues.

To evaluate the quality of PhraseStereo dataset, a comparative analysis across multiple scale factors was conducted. 
In the context of stereo image generation, the parameter scale factor, which we denote as $\beta$ controls the magnitude 
of disparity between the left and right views, effectively simulating the baseline distance between stereo cameras. 
A larger scale factor corresponds to a wider baseline. While this enhances depth cues, it also introduces several challenges. 
One of the most critical issues is the emergence and intensification of hallucinations, regions in the generated right image that 
do not correspond to any real content in the left image. These hallucinations typically arise when the model attempts to infer 
unseen areas, extrapolating geometry that is not visible in the input. However, due to the absence of corresponding real stereo 
data as ground-truth, i.e. stereo pairs where both views are captured from the same scene using a real stereo setup, 
we are unable to perform a direct analysis of hallucinations.

Instead, we focus on evaluating the perceptual and geometric quality of the generated stereo pairs.
As the scale factor increases, the model struggles to maintain both perceptual realism and geometric consistency, 
leading to distorted object shapes, misaligned edges, and even the invention of non-existent structures. 
This compromises the overall realism of the stereo pair and poses difficulties in preserving semantic coherence.
\begin{figure*}[h]
  \centering
  \includegraphics[page=1,width=0.76\textwidth]{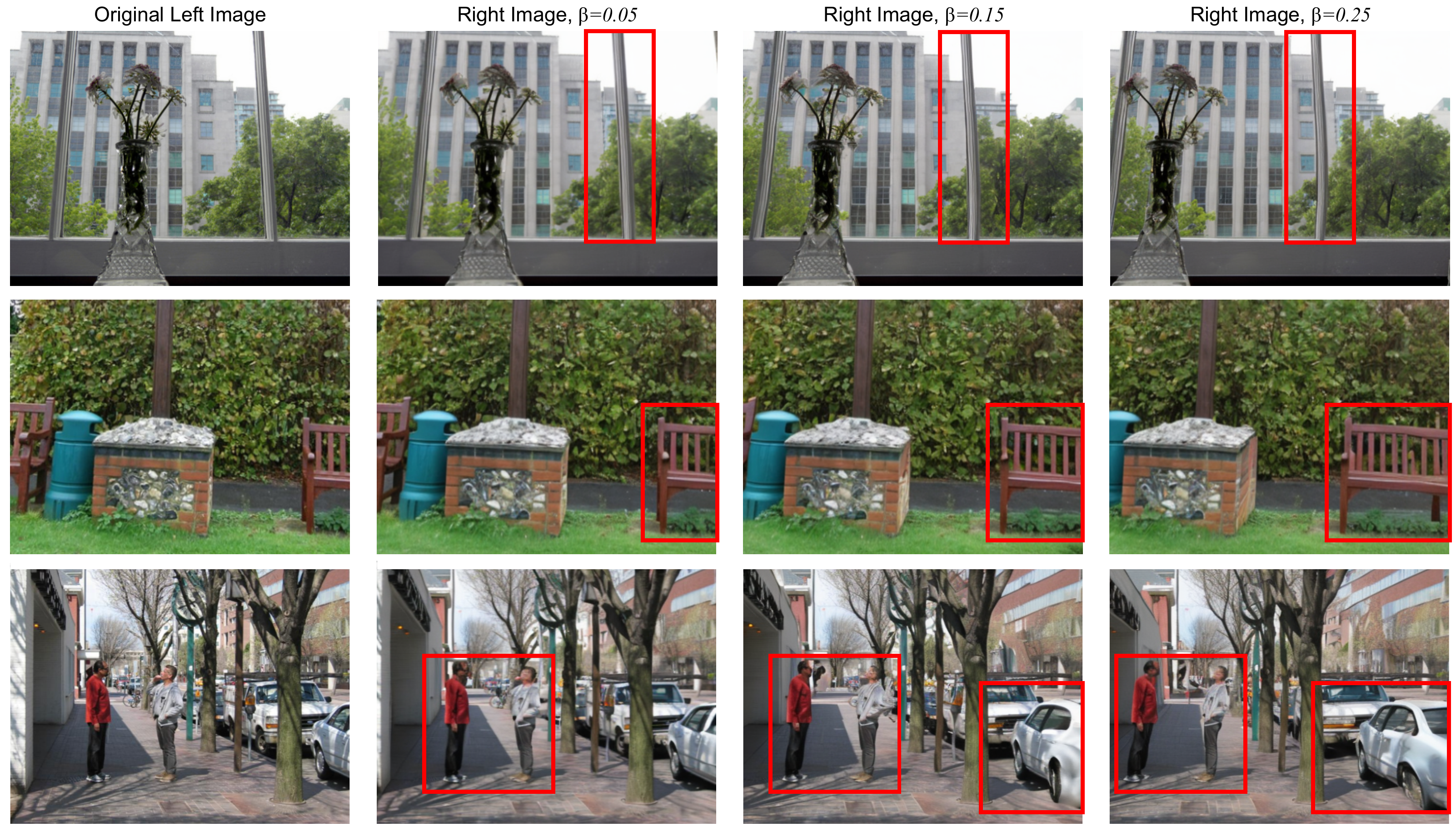}
  \caption{Qualitative analysis of right images in the PhraseStereo dataset across 
  different scale factors. From left to right, 
  each row shows the original left image~\cite{wu2020phrasecut}, the generated right images at 
  $\beta$ increasing. Red rectangles highlight 
  perceptual and geometric inconsistencies. PhraseStereo adopts $\beta = 0.15$ as a balanced configuration.}
  \vspace{-20pt}
  \label{fig:hallucinations}
\end{figure*}

For this analysis, we created three versions of the PhraseStereo dataset using scale factors $\beta \in \{0.05, 0.15, 0.25\}$. 
To assess the quality of the generated right images and address the issues described above, we adopted two complementary 
evaluation metrics: SSIM (Structural Similarity Index Measure)~\cite{wang2004image} and LPIPS (Learned Perceptual Image Patch Similarity)~\cite{zhang2018perceptual}, 
using the \textit{AlexNet}~\cite{krizhevsky2012imagenet} backbone network. These metrics were chosen for their relevance in the prior work GenStereo~\cite{qiao2025genstereo}.
\Cref{tab:scale_factor} presents the scores obtained across the different scale factors.\\
\begin{table}[h]
  \centering
  \begin{tabular}{@{}lcc@{}}
    Scale Factor & SSIM $\uparrow$& LPIPS $\downarrow$ \\
    \midrule
    0.05 & 0.679 & 0.227 \\
    \textbf{0.15 (final choice)} & 0.601 & 0.352\\
    0.25 & 0.571 & 0.412 \\
    \bottomrule
  \end{tabular}
  \caption{Quantitative analysis of generated right images across different GenStereo~\cite{qiao2025genstereo} scale factors using SSIM 
  and LPIPS metrics. Higher SSIM and lower LPIPS indicate better structural and perceptual similarity, respectively.}
  \vspace{-10pt}
  \label{tab:scale_factor}
\end{table}

Our experiments show that as the scale factor increases, both perceptual and structural similarity to the reference 
image decrease in performance. Specifically, with $\beta=0.05$, the generated images are perceptually close to the reference 
but exhibit minimal disparity, which limits their effectiveness for stereo learning. 
Higher scale factor, $\beta = 0.25$, introduces more pronounced disparity, but the generated images suffer from 
increased hallucinations and semantic drift, as reflected in the lower SSIM and higher LPIPS scores.
\Cref{fig:hallucinations} presents the qualitative results. In the first row, the model struggles in geometric consistency, 
with distortions more evident at $\beta = 0.25$. The second row highlights fine textures, where missing details appear 
at $\beta = 0.05$ and $\beta = 0.25$, while $\beta = 0.15$ preserves them well. The third row features strong semantic cues, 
making hallucinations more noticeable at higher scale factors. We selected $\beta = 0.15$ as the optimal scale factor for PhraseStereo, 
a balanced value determined through both quantitative metrics and qualitative observations. This configuration provides a realistic simulation of stereo camera systems, where moderate 
baselines are commonly employed to balance depth perception and image coherence.
This choice ensures that PhraseStereo provides stereo pairs that are both challenging and realistic, making the 
dataset suitable for training and evaluating models in stereo vision and referring expression semantic segmentation.

\section{Conclusion}
We presented PhraseStereo, a novel dataset for phrase grounding segmentation in stereo image pairs. By enabling models to 
leverage stereo geometry, PhraseStereo can facilitate more accurate segmentation of referred objects and regions. 
It provides a foundation for exploring multimodal architectures that integrate vision and language in a stereo context, 
paving the way for advances in both geometric reasoning and semantic understanding.

The performance on the PhraseStereo dataset, particularly in the generation of stereo image pairs, remains 
limited due to the introduction of hallucinated content when synthesizing right-view images from monocular 
images. These hallucinations and the geometry consistency underscore the current gap between synthetic stereo generation 
and the real stereo data. 
To address this, a promising future direction of research is the integration of a stereo disparity matching module into the pipeline. 
After generating the right view, a stereo matching model can evaluate its consistency with the left view and compute a 
disparity loss to guide training. This stereo supervision encourages geometrically consistent outputs and improves depth 
prediction, helping bridge the gap between synthetic and real stereo data.
{
    \small
    \bibliographystyle{ieeenat_fullname}
    \bibliography{main}
}

\end{document}